\documentclass{article}
\usepackage{spconf,amsmath,graphicx,hyperref}
\usepackage{xcolor, tabularx, booktabs, siunitx, bm, float}

\title{Acoustic Markers of Engagement \\in Children with Autism Spectrum Disorder and Their Caregivers}
\name{Young Suh Hong$^1$, Kaleen Shrestha$^2$, Austin Joel Dympep\thanks{This is work done while Austin Joel Dympep was an intern at the University of Southern California through the IUSSTF-Viterbi program.}$^3$, Maja Matarić$^2$}
\address{
  $^1$Department of Systems and Information Engineering, University of Virginia \\
  $^2$Computer Science Department, Viterbi School of Engineering, University of Southern California \\
  $^3$Department of Computer Science and Engineering, National Institute of Technology Meghalaya
}
\begin{document}
%
\maketitle
\begin{abstract}
\label{sec:abstract}
Characterizing user engagement is essential for developing effective human-computer interaction (HCI) applications, particularly for children with autism spectrum disorder (ASD). Because home-based HCI naturally involves caregivers as conversational facilitators, analyzing both child and caregiver vocalizations can provide critical insights into characterizing interaction dynamics. Using a dataset of home-based socially assistive robot (SAR) interventions, this work demonstrates how child and caregiver vocalization patterns can model engagement in HCI contexts. We employed linear mixed-effects models (LMMs) with child and caregiver acoustic data to identify acoustic features that significantly differentiated engagement from disengagement. The results demonstrated that children and caregivers share three significant acoustic markers of engagement: mean Alpha Ratio (AlphaRatio), mean Harmonics-to-Noise Ratio (HNR), and mean Spectral Slope 500-1500 (SlopeV500-1500). This reflects cleaner and smoother vocalizations during engagement across both groups. Furthermore, we evaluated a State $\times$ Group interaction model to compare the magnitudes of standardized state-dependent differences between children and caregivers. The results indicated that children exhibit significantly greater state-related variation than caregivers in three acoustic features: AlphaRatio, HNR, and SlopeV500-1500.

\end{abstract}
\section{Introduction}
\label{sec:intro}
Autism spectrum disorder (ASD) is a neurodevelopmental condition that influences how individuals communicate, behave, learn, and relate to others, affecting approximately 1 in 32 children in the United States \cite{shaw2025prevalence}. Previous research suggests that common human-computer interaction (HCI) technologies, such as screen-based applications, robots, and virtual reality, have a positive effect on the social and emotional development of children with ASD\cite{Yifu2024}. For HCI interventions to be effective, participants must remain actively engaged. Therefore, characterizing engagement provides valuable insights in developing transparent and interpretable HCI interventions.\\
\indent Vocal cues provide a non-invasive method for monitoring engagement in children with ASD. For example, Keerthana et al. \cite{keerthana2026speech} developed a speech-only engagement monitoring pipeline to guide personalized learning interventions, while Gupta et al. \cite{gupta12_wocci} demonstrated that vocal characteristics reflect engagement levels in children with ASD. Furthermore, in real-world and home-based environments, child engagement rarely occurs in isolation because caregivers actively participate as facilitators. Recognizing this dyadic interaction, prior work has incorporated caregiver audio alongside child vocalizations to model child engagement \cite{jain2020modeling, shi2022toward, chorianopoulou2017}.\\
\indent Despite the success of prior work in predicting child engagement using machine learning classifiers, existing approaches predominantly treat audio-based engagement detection as a classification task. As a result, limited work has examined which specific acoustic features systematically vary with a child's engagement state or how these state-dependent acoustic differences compare between children with ASD and their caregivers. Understanding these acoustic differences provides insight into the vocal correlates of child engagement and supports the development of interpretable HCI interventions for children with ASD.\\
\indent To address this gap, we conducted a systematic statistical evaluation of diarized speech segments from children with ASD and their caregivers during an in-home intervention with a Socially Assistive Robot (SAR). Using linear mixed-effects models (LMMs) with individual-level random intercepts, we first identified acoustic features associated with a child's engagement state separately for child and caregiver speech. We then constructed a State $\times$ Group interaction LMM to test whether the magnitude of state-dependent acoustic differences varied between child and caregiver speech. Together, these analyses characterized how the acoustic features of both children and caregivers varied with respect to a child's engagement state.
\section{Methodology}
\label{sec:methods}
\subsection{Data Collection}
We used data collected by our laboratory \cite{jain2020modeling, clabaugh, shi2022toward} during a month-long, in-home socially assistive robot (SAR) intervention with seven children diagnosed with ASD, ranging in age from three to seven years. During the intervention, participants played math games on a tablet while a squash-and-stretch SAR \cite{short2020sprite} provided encouragement verbally and through screen-based facial expressions and body movement. An average of three hours of video and audio recordings were collected from each child participant by a camera mounted on top of the tablet that captured the participant and their surrounding environment. Caregivers were present in the background for facilitating all session.

\subsection{Annotation}
\label{sec:methods:annotation}
We used binary engagement annotation labels that are part of the dataset \cite{jain2020modeling}. Each video frame was annotated as the child being either engaged (1) or disengaged (0). The annotation criteria for engagement were listening or responding to the robot's prompts, interacting with the tablet, or seeking guidance from the caregivers in the room. 

\subsection{Diarization Model Evaluation}
For speech diarization model evaluation, ground-truth speaker labels were established through manual annotation of 20\% of the data by two of the authors. We employed Pyannote \cite{bredin2023pyannote, plaquet2023powerset, pyannote2023speakerdiarization31} for speaker diarization, which achieved a mean diarization error rate (DER) of $29.9 \pm 9.3\%$ evaluated against the ground-truth label. The model was capable of detecting overlapping speech of multiple speakers.

\subsection{Data Preprocessing}
Following speaker diarization, we segmented the recordings into speaker-specific audio clips. We applied the timestamps generated by Pyannote to the original recordings to generate clips for each speaker, excluding segments containing overlapping speech. Each resulting clip contained speech from a single speaker and was subsequently used for acoustic feature extraction.\\
\indent We then mapped a child's engagement labels to both child and caregiver speech clips based on the annotations described in Section \ref{sec:methods:annotation}. When a speaker-specific clip spanned multiple child engagement annotations, we split the clip into distinct sub-clips corresponding to each engagement state to preserve the ground-truth labels. The distributions of child and caregiver audio clips across a child's engagement states are reported in Table \ref{tab:num_clips}.
\begin{table}[!htb]
    \caption{Number of audio clips by engagement state for children and parents}
    \label{tab:num_clips}
    \centering
    \vspace{2mm} 
    \begin{tabular}{cccc}
        \toprule
        \multicolumn{2}{c}{\textbf{Child}} & \multicolumn{2}{c}{\textbf{Parent}} \\
        \cmidrule(lr){1-2} \cmidrule(lr){3-4}
        Engaged & Disengaged & Engaged & Disengaged \\
        \midrule
        5,123 & 3,222 & 4,727 & 2,579 \\
        \bottomrule
    \end{tabular}
\end{table}

\subsection{Feature Selection}
\begin{table*}[!t]
\caption{Acoustic features used in this study}
    \label{tab:feature_list}
    \centering
    \vspace{2mm}
    \begin{tabularx}{\textwidth}{lXll}
    \toprule
       \textbf{Category} & \textbf{Feature Full Name} & \textbf{Name Used in Text} & \textbf{Abbreviation (if any)} \\
    \midrule
       Frequency related & F0semitoneFrom27.5Hz\_sma3nz\_amean & Pitch &\\
                         & jitterLocal\_sma3nz\_amean & Jitter &\\
                         & F1frequency\_sma3nz\_amean & F1Frequency &\\
                         & F2frequency\_sma3nz\_amean & F2Frequency &\\
                         & F3frequency\_sma3nz\_amean & F3Frequency &\\
                         & F1bandwidth\_sma3nz\_amean & F1Bandwith &\\
    \midrule
       Energy related    & shimmerLocaldB\_sma3nz\_amean & Shimmer &\\
                         & loudness\_sma3\_amean         & Loudness &\\
                         & HNRdBACF\_sma3nz\_amean       & HNR & Harmonics-to-Noise Ratio\\
    \midrule
       Spectral & alphaRatioV\_sma3nz\_amean         & AlphaRatio &\\
                          & hammarbergIndexV\_sma3nz\_amean    & Hammarberg &\\
                          & slopeV0-500\_sma3nz\_amean         & SlopeV0-500 &\\
                          & slopeV500-1500\_sma3nz\_amean      & SlopeV500-1500 &\\
                          & F1amplitudeLogRelF0\_sma3nz\_amean & F1LogRelF0 &\\
                          & F2amplitudeLogRelF0\_sma3nz\_amean & F2LogRelF0 &\\
                          & F3amplitudeLogRelF0\_sma3nz\_amean & F3LogRelF0 &\\
                          & logRelF0-H1-H2\_sma3nz\_amean      & H1H2LogRel &\\
                          & logRelF0-H1-A3\_sma3nz\_amean      & H1A3LogRel &\\
    \midrule
       Temporal           & loudnessPeaksPerSec                & LPR & Rate of loudness peaks\\
                          & MeanVoicedSegmentLengthSec         & VoicedLength &\\
                          & MeanUnvoicedSegmentLength          & UnvoicedLength &\\
                          & VoicedSegmentsPerSec               & PSR & Pseudo syllable rate\\
    \bottomrule
    \multicolumn{4}{@{}p{\textwidth}@{}}{\footnotesize \textit{Note.} The categorization is based on the standard division of eGeMAPS by \cite{eyben2016geneva}. The ``Name used in Text'' column is how each feature is named in this paper.}
    \end{tabularx}
\end{table*}
For acoustic features, we used the extended Geneva Minimalistic Acoustic Parameters Set (eGeMAPS), which contains 88 acoustic features \cite{eyben2016geneva}. The eGeMAPS features contain cepstral parameters that are known to be effective in modeling affective states \cite{eyben2016geneva}. To enhance the interpretability of our results, we followed \cite{han2025exploring} and selected a subset of 22 features related to the central tendencies (e.g., mean) of each acoustic dimension. The complete list of selected features is given in Table \ref{tab:feature_list}.

\section{Results and Discussion}
\label{sec:pagestyle}
\subsection{Separate Model}
We first fitted LMMs to examine the relationship between acoustic features and engagement states for child and caregiver within their groups (child or caregiver). For both models, we set the engagement state as the fixed effect and the subject (child or caregiver) as the random intercept. The LMM model is defined in Equation \eqref{eq:lmm_separate}, where $i$ denotes the subject, $j$ represents the observation index, $u_i \sim \mathcal{N}(0, \sigma_u^2)$ is the subject-specific random intercept, and $\epsilon_{i,j} \sim \mathcal{N}(0, \sigma_\epsilon^2)$ is the residual error. $State = 0$ is disengaged, and $State = 1$ is engaged. We call this the Separate Model.
\begin{equation}
\label{eq:lmm_separate}
Y_{i,j} = \beta_0 + \beta_1 \cdot State_{i,j} + u_{i} + \epsilon_{i,j}, State \in \{0, 1\}
\end{equation}
The eGeMAPS feature values were z-score standardized for each acoustic feature based on the mean and standard deviation across all individuals within their group before running LMMs. The standardization formula is defined in Equation \eqref{eq:std}, where $\bar{x}_{group}$ and $s_{group}$ denote the mean and standard deviation of the group to which subject $i$ belongs. This transformation enabled comparison across metrics that used different unit measurements and avoided potential LMM convergence problems. 
\begin{equation}
\label{eq:std}
z_{i,j} = \frac{x_{i,j} - \bar{x}_{\text{group}}}{s_{\text{group}}}, \quad group \in \{child, parent\}
\end{equation}

We adjusted the p-values using Benjamini-Hochberg FDR (threshold $<$ 0.05). We present the results in Tables \ref{tab:lmm_child} and \ref{tab:lmm_parent}. Only significant features are reported. The $\beta_1$ indicates the difference between engaged and disengaged states in each group's standardized units, as defined in Equation \eqref{eq:lmm_separate}. The LMM results show features that have significant differences between states for each group, respectively. \\
\begin{table}[htb]
    \caption{LMM results on engagement state and acoustic features for child diarized audio}
    \label{tab:lmm_child}
    \centering
    \vspace{2mm} 
        \begin{tabular}{lllll}
        \toprule
           \textbf{Feature name} & $\bm{\beta}_1$ & \textbf{SE} & \textbf{z} & \bm{$p_{\mathrm{FDR}}$} \\
        \midrule
            Shimmer&-0.132&0.023&-5.781&0.000***\\
        \midrule
            HNR&0.171&0.023&7.613&0.000***\\
        \midrule
            AlphaRatio&-0.214&0.022&-9.691&0.000***\\
        \midrule
            Hammarberg&0.081&0.022&3.608&0.001**\\
        \midrule
            SlopeV500-1500&-0.257&0.022&-11.767&0.000***\\
        \midrule
            H1H2LogRel&0.087&0.022&3.852&0.000***\\
        \midrule
            H1A3LogRel&0.109&0.022&5.001&0.000***\\
        \midrule
            LPR&-0.091&0.023&-4.013&0.000***\\
        \midrule
            VoicedLength&0.055&0.023&2.416&0.035*\\
        \midrule
            UnvoicedLength&0.058&0.023&2.517&0.029*\\
        \bottomrule
    \multicolumn{3}{@{}l}{\footnotesize \textit{Note.} *~$p < .05$, **~$p < .01$, ***~$p < .001$.}
    \end{tabular}
\end{table}

\begin{table}[htb]
    \caption{LMM results on engagement state and acoustic features for parent diarized audio}
    \label{tab:lmm_parent}
    \centering
    \vspace{2mm} 
        \begin{tabular}{lllll}
        \toprule
           \textbf{Feature name} & $\bm{\beta}_1$ & \textbf{SE} & \textbf{z} & \bm{$p_{\mathrm{FDR}}$} \\
        \midrule
            Loudness&-0.103	&0.022	&-4.645	&0.000***\\
        \midrule
            HNR&0.07	&0.024&	2.954&	0.014*\\
        \midrule
            AlphaRatio&-0.108	&0.024&	-4.563	&0.000***\\
        \midrule
            SlopeV500-1500&-0.099	&0.023&	-4.237&	0.000***\\
        \midrule
            PSR&-0.102&0.025&-4.158&0.000***\\
        \bottomrule
    \multicolumn{3}{@{}l}{\footnotesize \textit{Note.} *~$p < .05$, **~$p < .01$, ***~$p < .001$.}
    \end{tabular}
\end{table}

Engagement had a significant effect on 10 acoustic features for child diarized audio Table \ref{tab:lmm_child}), and for caregiver diarized audio, engagement was significantly associated with five acoustic features (Table \ref{tab:lmm_parent}). For both groups, the significant features were concentrated in energy-related, spectral, and temporal parameters. Three features--HNR, AlphaRatio, and SlopeV500-1500--were found to be significant in both groups. The coefficients showed shared directional alignment between child and caregiver; relative to disengaged states, engaged states showed an increase in HNR and a decrease in both AlphaRatio and SlopeV500-1500. \\
\indent The increase in HNR indicates higher harmonics compared to inharmonic noise during engaged state compared to disengaged state in both groups, reflecting a cleaner voice during the engaged state. The decrease in AlphaRatio indicates that high-frequency energy (1-5 kHz) decreased relative to low-frequency energy (50-1000 Hz). This shows that the overall spectral balance shifted towards lower frequencies (50-1000 Hz), which can reflect a smoother and softer voice compared to a sharp voice associated with higher frequencies. According to \cite{deboer2023acoustic}, SlopeV500-1500 reflects tension, where flatter mean spectral slope indicates a voice with more tension. Our results showed a steeper mean spectral slope in the engaged state, reflecting a less tense and smoother voice in the engaged state compared to the disengaged state. The overall analysis of the Separate Models showed that both child and caregiver produced cleaner, smoother, and higher quality vocalizations during the engaged state compared to the disengaged state. 

\subsection{Interaction Model}
While the results in Tables \ref{tab:lmm_child} and \ref{tab:lmm_parent} showed significant features within each group, they do not reflect the relationship between the two groups, namely whether the difference between the engagement states is significant between the two groups. Therefore, we fitted an LMM with the interaction between engagement state and group as fixed effects, and subject as random effects. We call this model the Interaction Model. The Interaction Model is defined in Equation \eqref{eq:lmm_interaction}, where $Group = 0$ is the child, and $Group = 1$ is the caregiver. $\beta_3$ represents the interaction between the engagement state and group, and the remaining notation follows Equation \eqref{eq:lmm_separate}.
\begin{equation}
\label{eq:lmm_interaction}
\begin{split}
Y_{i,j} = {} & \beta_0 + \beta_1 \cdot \text{State}_{i,j} + \beta_2 \cdot \text{Group}_i \\
& + \beta_3 \cdot (\text{State}_{i,j} \times \text{Group}_i) + u_i + \varepsilon_{i,j}
\end{split}
\end{equation}
\indent Considering the developmental differences between children and caregivers, who are adults, we used z-score standardization within each group before running LMM, following \eqref{eq:std}. Despite standardization, loudness did not converge and was excluded in further analysis. Similar to the Separate Models, we adjusted the p-values using Benjamini-Hochberg FDR (threshold $<$ 0.05). The results are presented in Table \ref{tab:lmm_interaction}, and only the significant features are reported. 
\begin{table}[htb]
\caption{LMM result on interaction between engagement state and group}
    \label{tab:lmm_interaction}
    \centering
    \vspace{2mm} 
        \begin{tabular}{lllll}
        \toprule
           \textbf{Feature name} & $\bm{\beta}_3$ & \textbf{SE} & \textbf{z} & \bm{$p_{\mathrm{FDR}}$} \\
        \midrule
            HNR&-0.101&0.033&-3.08&0.015*\\
        \midrule
            AlphaRatio&0.106&0.032&3.275&0.011*\\
        \midrule
            SlopeV500-1500&0.158&0.032&4.925&0.000***\\
        \bottomrule
    \multicolumn{3}{@{}l}{\footnotesize \textit{Note.} *~$p < .05$, **~$p < .01$, ***~$p < .001$.}
    \end{tabular}
\end{table}

As mentioned above, in Separate Models, both children ($\beta_1 = 0.171$, $p < 0.001$) and caregivers ($\beta_1 = 0.070$, $p < 0.05$) demonstrated significant positive standardized shifts in HNR during engaged vs. disengaged states. The Interaction Model confirmed a significant State $\times$ Group interaction ($\beta_3 = -0.101$, $p < 0.05$). This negative coefficient indicates that while both groups showed higher HNR during engagement, children showed a significantly stronger variation than caregivers.\\
\indent Similarly, both groups exhibited significant negative shifts during engagement in AlphaRatio ($\beta_1 = -0.214$, $p < 0.001$ for children; $\beta_1 = -0.108$, $p < 0.001$ for caregivers) and SlopeV500–1500 ($\beta_1 = -0.257$, $p < 0.001$ for children; $\beta_1 = -0.099$, $p < 0.001$ for caregivers). The Interaction Model also revealed significant State $\times$ Group interactions for both features ($\beta_3 = 0.106$, $p < 0.05$ for AlphaRatio; $\beta_3 = 0.158$, $p < 0.001$ for SlopeV500-1500). The positive interaction coefficients indicate that although both groups decreased in AlphaRatio and SlopeV500–1500 during engaged states, the magnitude of the drop was significantly stronger in children than in caregivers relative to their respective standardized distributions. Overall, children demonstrated a significantly greater magnitude of standardized shift across engagement states than caregivers specifically for HNR, AlphaRatio, and SlopeV500-1500, highlighting these features as prominent markers of engagement in children compared to caregivers. 

Across both models, no acoustic feature was found to be a strictly group-specific marker (reaching significance in only one group with a significant interaction). Similarly, no feature displayed a significant interaction effect without significant within-group variation between engagement states.

\section{Conclusion}
\label{sec:typestyle}
In this work, we examined acoustic differences between engagement and disengagement in children with ASD with respect to their caregivers' acoustic features during an in-home educational SAR intervention. Audio was diarized using Pyannote, and 22 eGeMAPS acoustic features were extracted. Separate Model LMMs that measured the difference between engagement states revealed 10 significant acoustic features associated with child engagement state in child audio and five in caregiver audio. Three features overlapped, indicating cleaner and smoother vocalizations during engagement across both groups. Furthermore, the Interaction Model (State $\times$ Group) LMM showed significantly stronger engagement-related shifts in HNR, AlphaRatio, and SlopeV500–1500 compared to caregivers. No acoustic feature served as an exclusively group-specific marker, and interaction effects occurred only alongside significant within-group state variation. These findings provide interpretable acoustic markers to guide the design of HCI interventions for children with ASD.


\vfill \clearpage \pagebreak

\section*{Acknowledgements}
\label{sec:acknowledgments}
Kaleen Shrestha is supported by the National Science Foundation CISE Graduate Fellowships under Grant No. 2313998.

\section*{Compliance with Ethical Standards}
\label{sec:compliance}
The data used in this work were collected previously with approval granted by the University of Southern California Institutional Review Board  (UP-16-00755).

\bibliographystyle{IEEEbib}
\bibliography{strings,refs}

\end{document}